\documentclass[letterpaper]{article} 
\usepackage{aaai23}  
\usepackage{times}  
\usepackage{helvet}  
\usepackage{courier}  
\usepackage[hyphens]{url}  
\usepackage{graphicx} 
\usepackage{natbib}  
\usepackage{caption} 
\usepackage{algorithm}
\usepackage{algorithmic}

\usepackage{amsmath}
\usepackage{amssymb}
\usepackage{bbm}
\usepackage{amsmath,bm}
\usepackage{xcolor}
\usepackage{multirow}

\usepackage{newfloat}
\usepackage{listings}
\DeclareCaptionStyle{ruled}{labelfont=normalfont,labelsep=colon,strut=off} 
\floatstyle{ruled}
\newfloat{listing}{tb}{lst}{}
\floatname{listing}{Listing}
\title{Efficient Image Captioning for Edge Devices}
\author{
	Ning Wang,~
	Jiangrong Xie,~
	Hang Luo,~
	Qinglin Cheng,~
	Jihao Wu,~
	Mingbo Jia,~
	Linlin Li \\
}
\affiliations{
    Huawei Inc.\\
    wn6149@mail.ustc.edu.cn,~
    xiexjr@foxmail.com,~
    \{lhjeremy,~qlincheng\}@outlook.com,~ \\
    \{wujihao,~jiamingbo,~lynn.lilinlin\}@huawei.com 
}

\usepackage{bibentry}

\begin{document}

\maketitle

\begin{abstract}
	Recent years have witnessed the rapid progress of image captioning.
	However, the demands for large memory storage and heavy computational burden prevent these captioning models from being deployed on mobile devices.
	The main obstacles lie in the heavyweight visual feature extractors (\emph{i.e.,} object detectors) and complicated cross-modal fusion networks.
	To this end, we propose LightCap, a lightweight image captioner for resource-limited devices.
	The core design is built on the recent CLIP model for efficient image captioning.
	To be specific, on the one hand, we leverage the CLIP model to extract the compact grid features without relying on the time-consuming object detectors.
	On the other hand, we transfer the image-text retrieval design of CLIP to image captioning scenarios by devising a novel visual concept extractor and a cross-modal modulator.
	%
	%
	We further optimize the cross-modal fusion model and parallel prediction heads via sequential and ensemble distillations.
	%
	%
	With the carefully designed architecture, our model merely contains 40M parameters, saving the model size by more than 75\% and the FLOPs by more than 98\% in comparison with the current state-of-the-art methods.  
	In spite of the low capacity, our model still exhibits state-of-the-art performance on prevalent datasets, \emph{e.g.,} 136.6 CIDEr on COCO Karpathy test split.
	Testing on the smartphone with only a single CPU, the proposed LightCap exhibits a fast inference speed of 188ms per image, which is ready for practical applications.
\end{abstract}

\section{Introduction}

%
%
Image captioning aims to automatically generate natural and readable sentences to describe the image contents, which provides a promising manner to help visually impaired people.
%
%
The recent decade has witnessed a surge of captioning algorithms, benefiting from the development of large-scale pre-training \cite{VLP,OSCAR,LEMON,SimVLM}, advanced representation learning \cite{VINVL,SOHO}, and modern cross-modal modeling \cite{E2E-VLP,OSCAR,ViTCap}.
In spite of the remarkable advances, current heavyweight captioning algorithms are not available to visually impaired people, who generally rely on low-resource devices such as portable phones to assist the daily life, instead of carrying on heavy computer servers with modern GPUs.
%
%
Designing computationally efficient and memory-friendly captioning methods is vital for practical applications but has been largely overlooked in the literature.

%
%
%
To achieve excellent performance, recent image captioners typically adopt deep object detectors as well as large cross-modal fusion networks.
For example, the recent VinVL and LEMON algorithms \cite{VINVL,LEMON} utilize a strong but heavyweight ResNeXt-152 based detection model and a base or large BERT model \cite{BERT}.
Some methods even scale the model size from base to huge to attain superior captioning performance \cite{LEMON}, but how to effectively reduce the model size for edge devices is rarely touched in these works.
These sophisticated image captioning models struggle to meet the real-time requirement of real-world applications, let alone the huge power consumption and memory storage.
It is therefore non-trivial to investigate how to design an efficient image captioner with smaller memory storage, faster inference speed, and satisfactory performance.
%
%
%

\begin{figure}
	\centering
	\includegraphics[width=8.1cm]{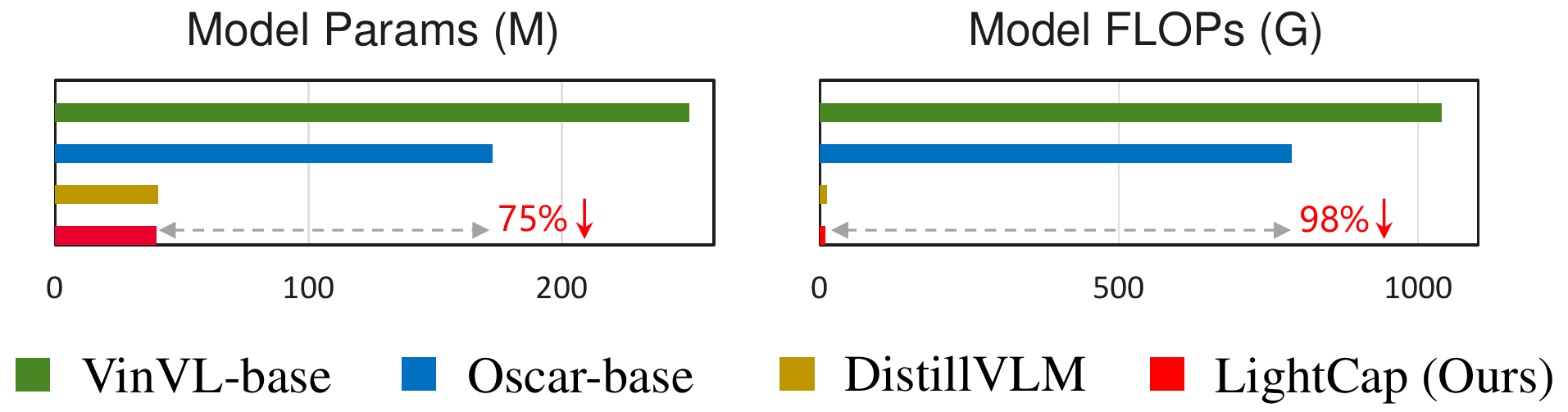}
	\caption{Compared to the state-of-the-art VinVL \cite{VINVL} and Oscar \cite{OSCAR}, our method saves more than 75\% parameters and 98\% FLOPs. Compared with the lightweight DistillVLM \cite{DistillVLM}, our method not only yields fewer parameters and FLOPs, but also outperforms it by a notable margin.
	}
	\label{fig:1}
\end{figure}

In this paper, we propose LightCap, a lightweight yet high-performance image captioning method for mobile devices.
Our core design is largely inspired by the recent CLIP method \cite{CLIP}.
CLIP is an impressive image-text retrieval model, which readily tells what objects exist in the image but fails to generate a description for the given image.
%
%
In this work, we investigate how to transfer such a strong cross-modal retrieval model to an image captioner, and meanwhile break the obstacles that hinder image captioners from being deployed on the mobile devices. 
%
%
The main obstacles that hinder image captioners from being deployed on mobile devices are their cross-modal fusion and image feature extraction models.
For visual representations, we leverage the efficient yet compact grid features from the CLIP without relying on time-consuming Region of Interest (ROI) features from sophisticated object detectors.
To unveil the potential of a capacity-limited model, we propose the following designs. 
%
(1) \emph{Visual concept extractor}. To take advantage of the cross-modal retrieval capability of CLIP, we train a region-based alignment model to retrieve the visual concepts from an off-the-shelf dictionary.
These visual concepts serve as the description hints of the image to facilitate caption generation.
%
%
(2) \emph{Cross-modal modulator}. Before being fed to the fusion model, the feature dimension of the CLIP feature is highly compressed (\emph{i.e.,} from 2048 to 312), which inevitably loses semantic representations.
To retain the valuable semantics, we propose a cross-modal modulator that takes the textual concepts as inputs to activate the informative feature channels of the CLIP model.
(3) \emph{Ensemble head}. We jointly optimize and distill an ensemble of head networks for collaborative prediction.
We disentangle the key parameters and share the rest weights of different heads for lightweight design.
%
%
Last but not least, for the cross-modal fusion model, instead of the widely-used $\text{BERT}_\text{base}$ \cite{BERT}, we chose the efficient TinyBERT \cite{TinyBERT} to fuse cross-modal features.
By virtue of our designed sequential knowledge distillations in both pre-training and fine-tuning stages and the ensemble distillations from multiple teachers, a TinyBERT almost matches the performance of the standard $\text{BERT}$.
%
%
%
%
%
%
%
%

By highly limiting the capacity of each component in our image captioner, the overall model merely contains 40M parameters and 9.8G FLOPs, saving the model size by more than 75\% and the FLOPs by more than 98\% compared to the current popular image captioning models (Figure~\ref{fig:1}).
Despite its low capacity, the proposed method still exhibits state-of-the-art performance on prevalent captioning datasets, \emph{e.g.,} 136.6 CIDEr on COCO Karpathy split \cite{MSCOCO}.
%
The model storage memory of LightCap is about 112MB, which is affordable on most mobile devices.
%
%
It merely costs about 188ms to process an image when testing the proposed LightCap on the mobile phone with only one CPU, which is readily ready for practical usage.
%
%

In summary, in this paper, we systematically show how to obtain a lightweight, efficient, and high-performance captioner by careful designs and training:
\begin{itemize}
	\item {\bf Model Design.} We propose a \emph{visual concept extractor} and a \emph{cross-modal modulator} to better exploit the cross-modal capability of the CLIP model for image captioning. 
	We further design a partially parameter-sharing \emph{ensemble head} for collaborative prediction.
	
	\item {\bf Model Training.} We present the \emph{sequential knowledge distillations} from pre-training to fine-tuning to distill the tiny model. We leverage the \emph{ensemble distillation} to better optimize the TinyBERT model and ensemble heads. 
\end{itemize}

\section{Related Work}

{\noindent \bf Image Captioning.} 
%
Image captioning methods generally contain a visual encoder to extract the image representations and a cross-modal fusion model to generate the caption. 
Previous methods \cite{AoANet,XLAN,BUTD,GET_AAAI,CAAG_AAAI,A2_AAAI,yang2022reformer} typically utilize the object detection methods such as Faster-RCNN \cite{FasterRCNN} to extract ROI features.
The recent VinVL method \cite{VINVL} shows that a strong visual feature extractor consistently improves the performance on image captioning.
To reduce the computational burden, MiniVLM \cite{miniVLM} designs a lightweight object detector using EfficientNet backbone \cite{Efficientnet}. DistillVLM \cite{DistillVLM} leverages knowledge distillation to acquire a thinner transformer architecture for vision-language tasks.
In contrast to the ROI features from object detectors, some cross-modal algorithms turn to the grid features for high efficiency, which are known as the detector-free approaches in the literature \cite{ViTCap,E2E-VLP,SimVLM,PureT_AAAI}.
%
%
%
%
Nevertheless, these models \cite{ViTCap,SimVLM,PureT_AAAI} still struggle to be deployed on edge devices.
%
%
Compared with them, our method leverages a light yet powerful CLIP model to extract the grid features. 
We further propose a concept extractor and a cross-modal modulator to unveil the cross-modal representation power of the CLIP.
Our approach outperforms previous efficient captioners such as MiniVLM \cite{miniVLM} and DistillVLM \cite{DistillVLM} with lower model capacity and faster inference speed, and is even comparable to the recent heavyweight captioners.

Recent works \cite{how_much_can_CLIP,universalCap} also take advantage of CLIP model for image captioning.
Nevertheless, they simply utilize the standard CLIP model to extract features or image tags.
In contrast, to reduce the model size, we train a lightweight region-level concept extractor as well as a feature modulator to better exploit the cross-modal characteristic of CLIP.

{\noindent \bf VL Pre-training.}
Vision-language (VL) pre-training aims to learn robust cross-modal representations to bridge the domain gap between vision and language signals \cite{empirical-study-VL}.
CLIP \cite{CLIP} and ALIGN \cite{ALIGN} align the VL representations via a light fusion manner (\emph{i.e,} dot-product) using the contrastive learning technique.
%
%
Nevertheless, their light fusion manner fails to conduct the cross-modal generation task such as image captioning.
In contrast, recent VL pre-training approaches \cite{VLP,UNITER,OSCAR,UNIMO,VINVL} adopt a relatively heavy transformer architecture \cite{Transformer} to fuse the VL representations, which are qualified to perform more VL downstream tasks.
Inspired by previous arts, our approach also involves VL pre-training to facilitate the downstream captioning task.
Differently, we do not employ the widely-adopted bidirectional masked language modeling, and shed light on the unidirectional language modeling to fully focus on the text generation task, \emph{e.g.}, image captioning.
Furthermore, similar to previous arts \cite{TinyBERT,mukherjee2020xtremedistil}, we adopt the sequential knowledge distillation (KD) to preserve the model representational capability within a tiny network.
Based on the general KD, we also investigate how to better leverage KD in the captioning task by introducing concept distillation to facilitate the modality alignment and ensemble distillation for multi-head optimization.

\section{Methodology}

In this section, we introduce the technical details of the proposed method. 
First, in Section \ref{model architecture}, we elaborate on the model design of each block.
Then, in Section \ref{model training}, we show the training details.
Finally, we exhibit the model distillation in both pre-training and fine-tuning stages in Section \ref{model distillation}.

\begin{figure}
	\centering
	\includegraphics[width=8.2cm]{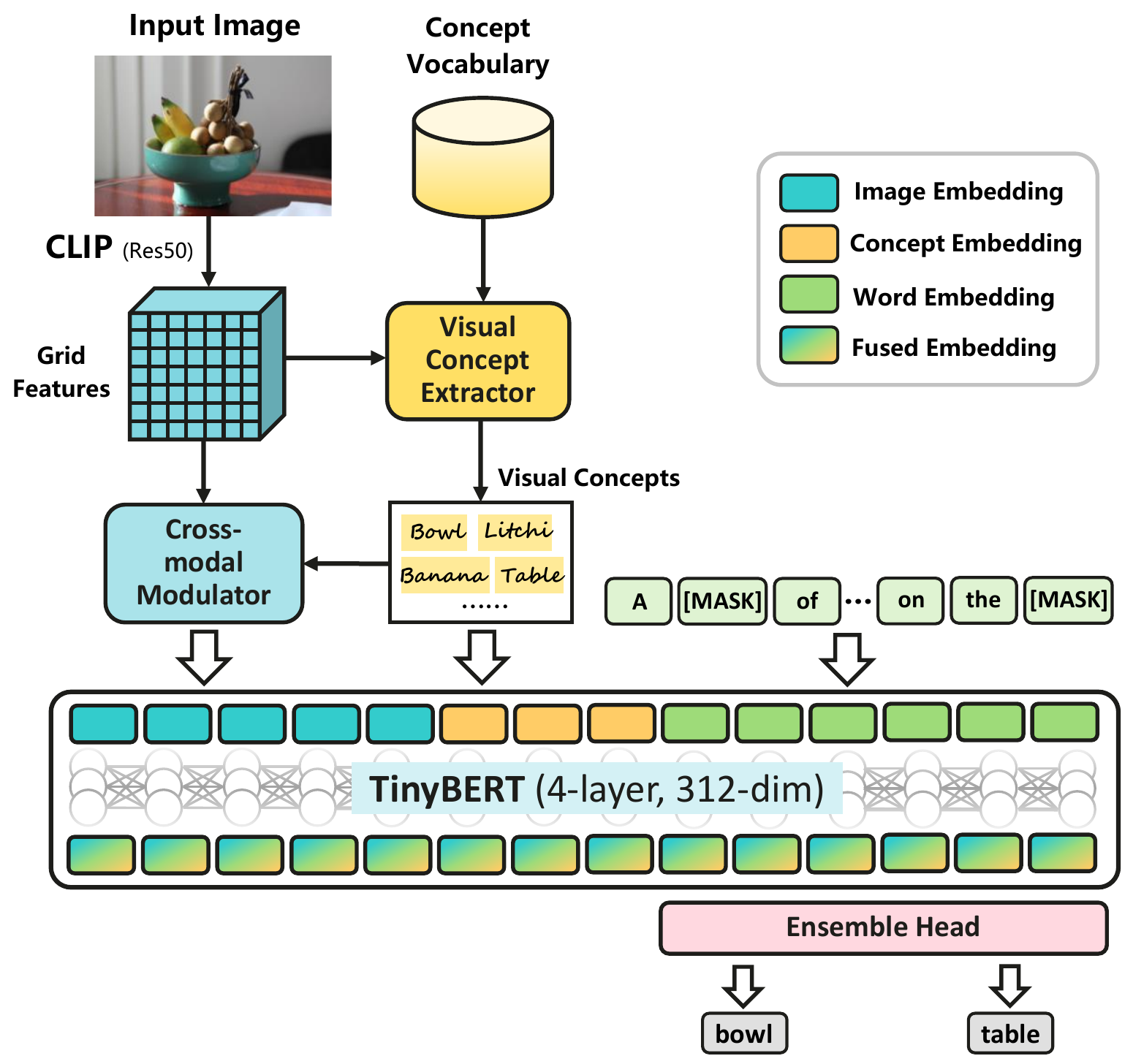}
	\caption{The overall framework of our LightCap. The input image is encoded to the grid visual feature via a ResNet-50 model. Then, we leverage a concept extractor to extract the visual concepts and a cross-modal modulator to reinforce the visual features. Finally, a TinyBERT fuses multi-modal embeddings and an ensemble head performs image captioning. }
	\label{fig:framework}
\end{figure}

\subsection{Model Architecture} \label{model architecture}
The overall framework is shown in Figure~\ref{fig:framework}. Our LightCap contains an image encoder to extract the visual representations, a concept extractor to retrieve the visual concepts from an off-the-shelf vocabulary, and a cross-modal modulator to enhance the visual representations with the textual (concept) information. Finally, we use a lightweight TinyBERT to fuse multi-modal representations and an ensemble head module to generate the image caption.

{\noindent \bf Image Encoder.}
Instead of extracting expensive ROI features from object detectors, we leverage the ResNet backbone \cite{ResNet} to acquire grid representations.
Specifically, we choose the recent CLIP model (ResNet-50 version) \cite{CLIP} due to (1) its impressive generalization capability, especially in the cross-modal domain; (2) its promising potential in extracting visual concepts from images, which is beneficial to the image captioning task.
%
%
CLIP model contains a visual encoder and a text encoder. 
In the visual encoder, after obtaining the image feature map, CLIP additionally learns a transformer block (\emph{i.e.}, attention pooler) to obtain the global image embedding. 
In our framework, to save the model capacity, we only utilize the ResNet-50 backbone in CLIP visual encoder \emph{without} the attention pooler to extract the visual features $ {\boldsymbol v} \in \mathbb{R}^{7 \times 7 \times 2048} $, which only involves 4.1G FLOPs.

{\noindent \bf Visual Concept Extractor.}
Intuitively, knowing the semantic concepts of the image is highly beneficial to image captioning.
Although CLIP model is ready for cross-modal retrieval, there still exist two issues.
First, CLIP relies on a heavy attention pooler to obtain the global image representation, which contains 14.8M parameters and is in conflict with our lightweight model design.
Second, CLIP model is pre-trained using global image features and thus is not effective enough in recognizing image regions.
To this end, we design and train an efficient region-based visual concept extractor on top of the CLIP feature.

\begin{figure*}
	\centering
	\includegraphics[width=17cm]{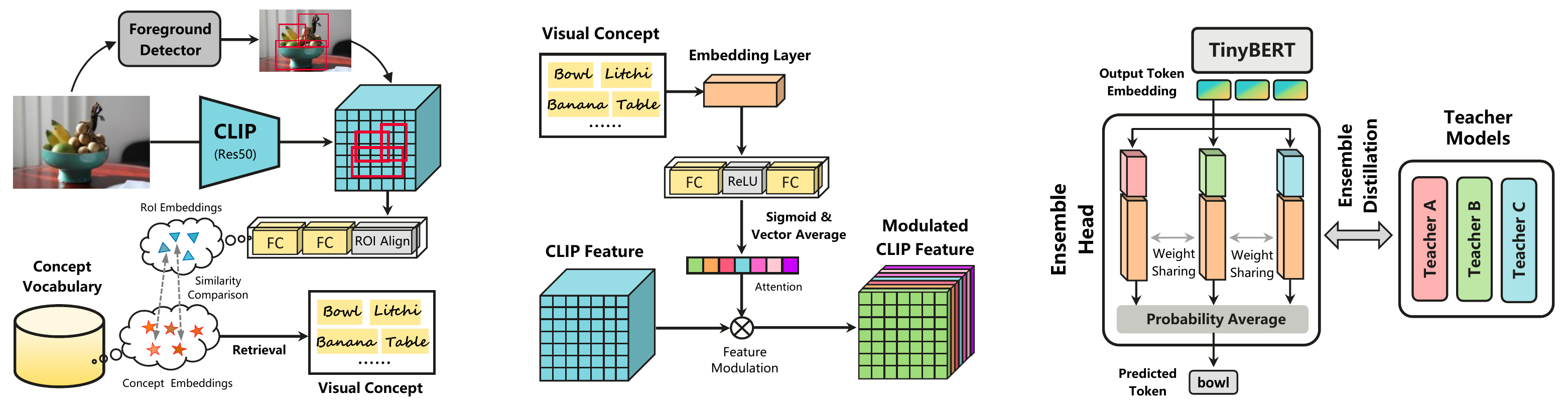}
	\caption{{\bf Left}: visual concept extractor block. {\bf Middle}: cross-modal modulator block. {\bf Right}: ensemble head block.}
	\label{fig:module} 
\end{figure*}

The overall architecture of the proposed visual concept extractor is shown in Figure~\ref{fig:module} (left).
First, we collect the common object categories from the Visual Genome dataset \cite{VG}, and form these category words using the description form $\texttt{a}$ $\texttt{photo}$ $\texttt{of}$ $\texttt{[object]}$.
We take advantage of the CLIP text encoder to extract the textual embeddings of these descriptions to form an off-the-shelf vocabulary.
Note that this vocabulary contains textual embeddings instead of the raw words to avoid unnecessary computations in the captioning stage.
Then, we train an efficient foreground-background object detector without knowing object classes.
This detector is designed to roughly predict the foreground bounding boxes, whose architecture is tiny YOLOv5n \cite{YOLOV5} with only 1.9M parameters.
After obtaining the object proposals, we employ ROI-Align \cite{MaskRCNN} to pool the region embeddings.
These ROI embeddings are further processed by two linear blocks to align with the concept embeddings in the aforementioned vocabulary.
To train this concept extractor, we freeze the CLIP ResNet-50 parameters and only train two linear layers using the standard contrastive loss in CLIP.

In summary, compared to the original CLIP, we transfer it from global image-text retrieval to region-level content retrieval.  
In the image captioning stage, for each foreground proposal, the object category with the highest similarity score is assigned as its label.
All the retrieved labels are assembled to form the visual concept of the image.

{\noindent \bf Cross-modal Modulator.}
ResNet-50 backbone yields the feature map with a high channel dimension of 2048, which requires to be highly compressed before multi-modal fusion.
It has been well recognized that different feature channels contain certain semantics.
After extracting the visual concepts that reside in the image, we propose to utilize these textual hints to promote the visual representations.
Specifically, we train a modulator that receives the concept tokens to activate the informative channels of the CLIP feature.
As shown in Figure~\ref{fig:module} (middle), this cross-modal modulator contains an embedding layer to embed the concept words, two fully-connected layers with a non-linear ReLU function to project the word embeddings, and a Sigmoid function to restrict the output weight.
Finally, we average the output weights of all the concepts to obtain the final channel activation weight ${\bf w} \in \mathbb{R}^{1 \times 1 \times 2048} $, which is applied to the raw CLIP feature ${\boldsymbol v}$ to reweigh the channel importance via ${\boldsymbol v}^{\diamond} = {\bf w}\otimes{\boldsymbol v} $, where $\otimes$ denotes the channel-wise multiplication, and ${\boldsymbol v}^{\diamond}$ is the modulated CLIP feature.

{\noindent \bf Multi-modal Fusion Module.} The proposed method adopts $\text{TinyBERT}_{4}$ \cite{TinyBERT} as the cross-modal fusion module, which is extremely shallow consisting of only 4 transformer blocks and a hidden size of 312.
%

%
Following previous arts \cite{OSCAR,VINVL}, we apply the \texttt{seq2seq} attention mask to generate the caption token in an auto-regressive
way.
Our TinyBERT takes as input the concatenation of the modulated image features ${\boldsymbol v}^{\diamond}$ and visual concept embeddings ${\boldsymbol c}$, and starts the caption generation by appending a mask token $[\texttt{MASK}]$ to the inputs. Then, the previous $[\texttt{MASK}]$ is replaced by the predicted token, and a new $[\texttt{MASK}]$ is appended to generate the next word.
The words are predicted one by one until the TinyBERT outputs the $[\texttt{STOP}]$ token.

{\noindent \bf Ensemble Head Module.} 
Multi-model ensemble is an intuitive way to improve the performance, but will greatly increase the model size.
In this work, we propose a parameter-efficient ensemble head to predict the token.
%
%
The ensemble head contains three branches to parallelly tackle the word embeddings, as shown in Figure~\ref{fig:module} (right).
We recognize that the parameter burden of head network mainly resides in the word embedding layer, whose shape is $312\times 30522$ (dictionary size).
To reduce the storage room, word embedding layers in different branches share the model weights, while the lightweight project layers (shape: $312\times 312$) before the word embedding layer are individually optimized for diversity.
These parallel head networks are individually distilled by different teacher networks to further enhance the prediction diversity, which will be discussed in the next section.

\subsection{Model Training} \label{model training}
{\noindent \bf Pre-training Stage.} Most VL pre-training methods \cite{LXMERT,UNITER,OSCAR,VINVL} utilize the popular masked language modeling (MLM) loss to pre-train the cross-modal fusion model.
Since our work focuses on the image captioning scenario, we do not apply the bi-directional modeling manner and choose the sequence-to-sequence MLM to facilitate the text generation.
To simulate the uni-directional generation process, the self-attention mask is constrained such that the caption token can only attend to the previous tokens.
To be specific, we randomly mask 15\% of the caption tokens following BERT and replace them with the special token $[\texttt{MASK}]$.
The fusion model takes the Image-Concept-Caption triple $(\boldsymbol{v}^{\diamond}, \boldsymbol{c}, \boldsymbol{x})$ from the dataset $\cal{D}$ as input, where $\boldsymbol{x}=\{{x}_1,\cdots, {x}_{T}\}$ are the masked input tokens.
The training objective is to reconstruct the masked token ${x}_{t}$ based on the previous tokens ($ \boldsymbol{x}_{<t} $), concepts ($\boldsymbol{c}$), and image features ($\boldsymbol{v}^{\diamond}$) by minimizing the following negative log-likelihood:
\begin{equation}\label{caption loss}
	{\cal L}_{\text{caption}} = - {\mathbb E}_{ (\boldsymbol{v}^{\diamond}, \boldsymbol{c}, \boldsymbol{x}) \in {\cal D}} \Big[ \sum_{t} \text{log}P(x_t| {\boldsymbol v}^{\diamond}, {\boldsymbol c}, {\boldsymbol x}_{<t}) \Big].
\end{equation}


Recent works \cite{OSCAR,VINVL} observe that image detection tags are qualified to serve as the anchor points to facilitate the multi-modal representation alignment.
Inspired by these methods \cite{OSCAR,VINVL}, we treat our retrieved visual concepts as the anchors to form the modality contrastive loss. 
To be specific, we ``pollute'' the image concept by replacing it with probability 50\% with a different concept from the dataset $\cal{D}$. The potentially polluted image concept is denoted by $\boldsymbol{c}^{\star}$.
We use a binary classifier $f(\cdot)$ on the top of the TinyBERT $[\texttt{CLS}]$ embedding to judge whether the triple $ ( {\boldsymbol v}^{\diamond}, {\boldsymbol c}^{\star}, {\boldsymbol x}) $ is polluted ($y=0$) or not ($y=1$). 
This concept contrastive loss $ {\cal L}_{\text{concept}} $ is defined as follows:
\begin{equation}\label{contrastive loss}
{\cal L}_{\text{concept}} = - {\mathbb E}_{ (\boldsymbol{v}^{\diamond}, \boldsymbol{c}^{\star}, \boldsymbol{x}) \in {\cal D}} \big[ \text{log}P(y|f( {\boldsymbol v}^{\diamond}, {\boldsymbol c}^{\star}, {\boldsymbol x})) \big].
\end{equation}

The aforementioned two losses are equally combined to form the final training objective in the pre-training stage: $ {\cal L}_{\text{pre-train}} =  {\cal L}_{\text{caption}}  +  {\cal L}_{\text{concept}} $.

{\noindent \bf Fine-tuning Stage.} 
After model pre-training on the noisy pre-training data, our LightCap model is further fine-tuned on the well-annotated captioning dataset such as COCO.  
In the fine-tuning stage, we do not adopt the contrastive loss and only utilize Eq.~(\ref{caption loss}) as the training objective to fully concentrate on the image captioning scenario.


\subsection{Knowledge Distillation} \label{model distillation}
%
We further adopt knowledge distillation (KD) to remedy the performance drop caused by the limited model capacity.
We train teacher networks with the architecture of $\text{BERT}_\text{base}$, and then sequentially distill the student model. %

{\noindent \bf KD in Pre-training Stage.}
In the pre-training stage, we first encourage the student model to mimic the transformer attentions and hidden state representations of its teacher:
\begin{equation}\label{KD_attention}
	\small
\begin{aligned}
{\cal L}_{\text{KD-1}} &= {\cal L}_{\text{atten}}^{\text{KD}} + {\cal L}_{\text{hidden}}^{\text{KD}} \\ &= \frac{1}{h}\sum_{i=1}^{h} \texttt{MSE}\left( {\bf A}_{i}^{\text{S} } , {\bf A}_{i}^{\text{T} } \right) + \frac{1}{l} \sum_{j=1}^{l} \texttt{MSE} \left( {\bf H}_{j}^{\text{S}} {\bf W}, {\bf H}_{3\times j}^{\text{T}} \right),
\end{aligned}
\end{equation}
where $\texttt{MSE}(\cdot, \cdot)$ denotes the mean-squared loss;
$ {\bf A}_{i}^{\text{S}} $ and $ {\bf A}_{i}^{\text{T}} $ are the attentions from the $i$-th head of the student model and teacher model, respectively;
$ {\bf H}_{j}^{\text{S}} $ and $ {\bf H}_{3\times j}^{\text{T}} $ denote the $j$-th and $(3\times j)$-th layer's hidden state representations from the student and teacher models, respectively (we empirically adopt this setting since the teacher model is 3 times deeper than the student model); 
$ {\bf W} $ is an $ 1\times 1$ linear block to facilitate the student model to match its teacher's feature dimension for hidden state distillation.

After the attention and hidden representation distillations, we further perform the second-stage KD, \emph{i.e.,} prediction-level distillation $ {\cal L}_{\text{KD-2}} $ as follows:
\begin{equation}\label{KD_prediction}
	\small
{\cal L}_{\text{KD-2}} = {\cal L}_{\text{caption}}^{\text{KD}} + {\cal L}_{\text{concept}}^{\text{KD}}= \texttt{CE}\left({\bf z}^{\text{S}}/\tau, {\bf z}^{\text{T}}/\tau \right) +  \texttt{CE}\left({\bf y}^{\text{S}}/\tau, {\bf y}^{\text{T}}/\tau \right),
\end{equation}
where $\texttt{CE}(\cdot, \cdot)$ denotes the cross-entropy loss, ${\bf z}^{\text{S}}$ and ${\bf z}^{\text{T}}$ denote the soft predictions of the tokens of the student and teacher; ${\bf y}^{\text{S}}$ and ${\bf y}^{\text{T}}$ are the ``pollution'' probability of the visual concepts of the student and teacher; $\tau$ refers to the temperature in KD.
In this distillation stage, the student model not only mimics the captioning capability (\emph{i.e.,} token prediction probability) of the teacher via ${\cal L}_{\text{caption}}^{\text{KD}}$, but also preserves the cross-modal alignment capability (\emph{i.e.,} concept ``pollution'' probability) via ${\cal L}_{\text{concept}}^{\text{KD}}$.

{\noindent \bf KD in Fine-tuning Stage.} 
%
%
In the fine-tuning stage, we also first conduct knowledge distillation on attention weights and hidden states as in Eq.~(\ref{KD_attention}), and then conduct knowledge distillation on the output probability. 
However, the model fine-tuning stage only involves a simple captioning constraint without the concept contrastive learning. 
Consequently, we merely force the student to mimic the token prediction of its teacher via $ {\cal L}_{\text{caption}}^{\text{KD}} = \texttt{CE}\left({\bf z}^{\text{S}}/\tau, {\bf z}^{\text{T}}/\tau \right) $.

{\noindent \bf Ensemble KD.} Actually, instead of adopting a single head, we construct the ensemble head with three parallel branches.
We train three teacher models with different model initializations.
These teachers jointly distill different branches of the ensemble head model, as shown in Figure~\ref{fig:module} (right).

\section{Experiments}

\subsection{Datasets and Metrics}

{\bf \noindent Pre-training Datasets.} In the experiments, we collect the image-text pairs from Google Conceptual Captions (CC3M) \cite{CC3M}, SBU Captions \cite{SBU}, OpenImages \cite{OpenImage}, and MS-COCO \cite{MSCOCO} to form the pre-training data.
In total, our pre-training corpus consists of about 5.8M image-text pairs.

{\bf \noindent Evaluation Datasets and Metrics.} We evaluate the proposed method on the COCO caption of Karpathy split \cite{MSCOCO} and nocaps validation dataset \cite{nocaps}.
%
%
%
To evaluate the quality of the generated captions, we use standard metrics in the image captioning task, including BLEU@4 \cite{BLEU}, METEOR \cite{METEOR}, CIDEr \cite{CIDER}, and SPICE \cite{SPICE}. 
In the captioning stage, beam search (beam size = $5$) is adopted in all experiments and the maximum generation length is restricted to $20$ words.
%

\subsection{Implementation Details}

{\bf \noindent Visual Encoder.} We take the ResNet-50 backbone from the CLIP model \cite{CLIP} as the visual feature extractor, whose parameters are frozen in both pre-training and fine-tuning stages. The input image resolution is $ 224 \times 224$.
%
%

{\bf \noindent Visual Concept Extractor.} We follow the tiny YOLOv5n \cite{YOLOV5} and its default settings to train a binary (foreground-background) object detector. 
This tiny detector is trained using Visual Genome dataset \cite{VG}, where all the object bounding boxes are treated as the foreground annotations.
After obtaining the foreground object detector, we train the alignment module using the region-level CLIP features and textual embeddings from the Visual Genome dataset.
This alignment module only contains two linear blocks ($2048 \times 1024$ and $1024 \times 1024$) and is trained for $60$ epochs with a learning rate of $1\times10^{-5}$.
%

{\bf \noindent Cross-modal Modulator.} The cross-modal modulator contains two sequential linear blocks with sizes of $312 \times 39$ and $39 \times 2048$.
The token embedding layer in this modulator shares weights with the embedding layer in TinyBERT.
%

{\bf \noindent Cross-modal Fusion Model.}
For the TinyBERT, we initialize it with the pre-trained weights \cite{TinyBERT}.
%
The visual concepts, as well as the caption words, are tokenized and projected via an embedding layer before being fed to the TinyBERT.
The modulated visual embeddings are compressed via the $1 \times 1$ linear block to match the TinyBERT's embedding dimension. 
%
%
In the pre-training stage, the fusion model is trained $1.0$M steps with a learning rate of $5\times10^{-5}$ and batch size of $512$.
%
%
%
In the fine-tuning stage, the fusion model is trained $120$ epochs with a learning rate of $3\times10^{-5}$. 
%
%
Except for the TinyBERT, we also train large fusion models $\text{BERT}_{\text{base}}$ \cite{BERT} following the above steps.
%
%

\subsection{Ablation Study}

\setlength{\tabcolsep}{2pt}
\begin{table}
	\scriptsize
	\begin{center}
		\begin{tabular*}{8.2 cm} {@{\extracolsep{\fill}}lcccc|cccc}
			\hline
			~   &Student   &Pre-training & Concept  & Modulator &B@4 &M &C &S \\
			\hline
			~   &\checkmark & & & &32.1 &26.9 &103.6 &19.9  \\
			~   &\checkmark &\checkmark & & &33.6 &27.7 &111.6 &20.8  \\
			~   &\checkmark &\checkmark &\checkmark &   &34.3 &28.3 &115.0 &21.3  \\
			~   &\checkmark &\checkmark &\checkmark &\checkmark &34.9 &28.9 &116.8 &21.9 \\
			\hline
		\end{tabular*}
		
		\vspace{+0.05in}
		
		\begin{tabular*}{8.2 cm} {@{\extracolsep{\fill}}lcccc|cccc}
			\hline
			~   &Teacher  &Pre-training  & Concept & Modulator &B@4 &M &C &S \\
			\hline
			~   &\checkmark  & & & &34.2 &28.3 &113.8 &21.2 \\
			~   &\checkmark &\checkmark & & &36.2 &29.0 &120.5 &22.1  \\
			~   &\checkmark &\checkmark  &\checkmark & &37.0 &29.6 &124.2 &23.5 \\
			~   &\checkmark &\checkmark  &\checkmark &\checkmark &37.5 &29.9 &126.3 &24.3 \\
			\hline
		\end{tabular*} 
		\caption{Ablative study of the proposed LightCap. To better investigate the performance of each component, the student model does not employ any knowledge distillation and uses a single head model. The evaluation metrics are BLEU@4 (B@4), METEOR (M), CIDEr (C), and SPICE (S) scores on the COCO-caption Karpathy test split \cite{MSCOCO}. } \label{table:ablation}	
	\end{center}
\end{table}


{\bf \noindent Model Pre-training.} It has been well recognized that model pre-training on large-scale image-text corpus benefits the image captioning. 
As shown in Table~\ref{table:ablation}, for the student model with limited capacity, model pre-training significantly improves the performance by 8.0 CIDEr score.

{\bf \noindent Visual Concept Extractor.}
The proposed visual concept extractor provides valuable clues for image captioning via an efficient image-text retrieval manner.
As shown in Table~\ref{table:ablation}, for the student model, the visual concept extractor improves the captioning performance by 3.4 CIDEr score on the COCO dataset.
This mechanism also improves the strong teacher model by 3.7 CIDEr score.

{\bf \noindent Cross-modal Modulator.} The cross-modal modulator takes advantage of the retrieved visual concepts to modulate the raw CLIP features.
%
%
As shown in Table~\ref{table:ablation}, based on the student model with a visual concept extractor, the proposed cross-modal modulator further improves the captioning performance by 1.8 CIDEr score.
This tiny block promotes the strong teacher model by 2.1 CIDEr score.  
%

{\bf \noindent Sequential Model Distillation.} In Table~\ref{table:distill}, we ablate the model knowledge distillation (KD) techniques in our approach.
First, we investigate KD in the pre-training stage in Table~\ref{table:distill} (top). 
In these experiments, we only adopt the standard cross-entropy optimization without any KD in the fine-tuning stage.
%
%
In the pre-training stage, the ``attention \& representation distillation'' improves 0.8 CIDEr score, and the distillation of output token probability improves 2.0 CIDEr score.
%
%
Considering the characteristic of cross-modal training, we further propose to distill the soft prediction of the anchor words (\emph{i.e.,} visual concepts), which brings an additional 1.2 CIDEr gain.
This indicates the concept distillation facilitates the cross-modal alignment.

\setlength{\tabcolsep}{2pt}
\begin{table}
	\scriptsize
	\begin{center}
		\begin{tabular*}{8.4 cm} {@{\extracolsep{\fill}}lccc|cc|c|cc}
			\hline
			~ & \multicolumn{3}{c|}{Pre-training Stage}  & \multicolumn{2}{c|}{Fine-tuning Stage} &Ensemble & \multicolumn{2}{c}{COCO test} \\
			~   &Atten$\&$Rep  & Caption &Concept  &Atten$\&$Rep &Caption &Distill  &B@4 &C \\
			\hline
			~   & & & & &   &  &34.9 &116.8 \\
			\hline
			~   &\checkmark & & & &    & &35.2 &117.6 \\
			~   &\checkmark &\checkmark & & &   & &35.6 &119.6 \\
			~   &\checkmark &\checkmark &\checkmark & & & &36.2   &120.8 \\
			\hline
			~   &\checkmark &\checkmark &\checkmark &\checkmark & & &36.4  &121.9 \\
			~   &\checkmark &\checkmark &\checkmark & &\checkmark & &36.8  &123.4  \\
			~   &\checkmark &\checkmark &\checkmark &\checkmark &\checkmark & &37.1  &124.1 \\
			\hline
			~   &\checkmark &\checkmark &\checkmark &\checkmark &\checkmark &\checkmark  &37.4  &125.8 \\
			
			\hline
		\end{tabular*}
		\caption{Ablative study of the proposed LightCap method using different distillation techniques. ``Atten\&Rep'', ``Caption'', and ``Concept'' denote the knowledge distillations on attention weight and hidden representation, token probability, and concept probability, respectively. Finally, we adopt the ensemble head block and leverage the ensemble distillation to optimize the overall model.} \label{table:distill}	
	\end{center}
\end{table}

Next, we investigate KD in the model fine-tuning stage.
As shown in Table~\ref{table:distill}, based on the distilled fusion model from the pre-training stage, in the fine-tuning stage, ``attention \& representation distillation" and ``output token distillation" further improve 1.1 CIDEr and 2.6 CIDEr, respectively.
Combining the above KD techniques achieves the best result of 3.3 CIDEr gain.
Finally, by virtue of the model distillation in both pre-training and fine-tuning, our lightweight student model achieves a promising captioning performance of 37.1 BLEU@4 and 124.1 CIDEr, and even matches the strong teacher model (\emph{i.e.,} 37.5 BLUE@4 and 126.3 CIDEr in Table~\ref{table:ablation}).

{\bf \noindent Ensemble Model Distillation.} The above experiments are based on the single head setting.
Actually, our model adopts the ensemble head for superior performance.
To encourage the prediction diversity, we prepare three teachers to individually distill these heads.
As shown in Table~\ref{table:distill}, ensemble head module and ensemble KD improve 1.7 CIDEr.

\subsection{Inference on the Mobile Device}\label{experiment on mobile devices}

Table~\ref{table:capacity of LightCap} exhibits the model FLOPs and parameters of each block in the LightCap.
%
%
%
%
%
Note that the ResNet50 backbone in CLIP adopts the half-precision model training and thus the model storage size of the visual encoder is 56.5MB.
Overall, our LightCap consumes a total storage space of 112.5MB, which is affordable for most mobile devices.

\setlength{\tabcolsep}{2pt}
\begin{table}
	\scriptsize
	\begin{center}
		\begin{tabular*}{8.5 cm} {@{\extracolsep{\fill}}l|cccc|c}
			\hline
			 &Img. Encoder &Concept Extractor  & Modulator  &Fusion  &Total\\
			\hline
			   Model & ResNet50 &YOLOv5n &2 FC  &$ {\text{TinyBERT}}_{\text{4}} $ & - \\
			   Params (M) &23.5M  &1.9M  &9.4$\times 10^{-2}$M  &14.5M  &39.9M    \\
			   Size (MB) &56.5MB  & 7.6MB  &0.4MB  &58.0MB  &112.5MB    \\
			   FLOPs (G) &4.1G &4.5G  &1.9$\times 10^{-4}$G  &1.2G &9.8G     \\
			\hline
		\end{tabular*}
		\caption{Illustration of model details including number of parameters (in M), model size (in MB), and computational complexity (FLOPs, in G) of the proposed LightCap.} \label{table:capacity of LightCap}	
	\end{center}
\end{table}

%
%
Then, we test the inference latency of LightCap model on Huawei P40 smartphone with a Kirin 990 chip.
%
%
To purely investigate the model inference speed, we set the beam search size to 1.
It merely takes about 188ms for our light model to process a single image on the CPU from mobile devices, which meets the real-world efficiency requirements.
%

\subsection{State-of-the-art Comparison}

{\bf \noindent Comparison on Model Size and Efficiency.}
%
%
In Table~\ref{table:params and FLOP}, we compare our LightCap with the state-of-the-art captioning methods in terms of model size and inference efficiency in FLOPs.
%
%
Most existing pre-training methods such as VLP \cite{VLP}, Oscar \cite{OSCAR}, and UNIMO \cite{UNIMO} use the Faster R-CNN as the feature extractor and a $\text{BERT}_{\text{base}}$ as the fusion model, yielding about 173M parameters and about 800G FLOPs.
It is worth noting that the current performance leaders such as VinVL \cite{VINVL} and LEMON \cite{LEMON} contain a huge FLOPs of more than 1000G.
As illustrated in Section~\ref{experiment on mobile devices}, the overall FLOPs of our LightCap is only 9.8G.
Consequently, compared with the recent popular image captioners, our LightCap saves more than 98\% of the FLOPs. 
%

\setlength{\tabcolsep}{2pt}
\begin{table}[h]
	\scriptsize
	\begin{center}
		\begin{tabular*}{8.6 cm} {@{\extracolsep{\fill}}l|ccc|ccc}
			\hline
			\multirow{2}{*}{Method} & \multicolumn{3}{c|}{Image Encoder}  & \multicolumn{3}{c}{Fusion Model}  \\
			 & Model & Params  &FLOPs  &Model  &Params  &FLOPs   \\
			\hline
			
			$\text{VinVL}_\text{B}$, $\text{LEMON}_\text{B}$ & $\text{ResNeXt}_\text{152}$  &141.2M &1017.2G & $\text{BERT}_\text{base}$    &109M  &22.5G   \\
			
			$\text{Oscar}_\text{B}$, $\text{VLP}_\text{B}$  & $\text{F-RCNN}_\text{101}$  &63.8M  &767.0G  & $\text{BERT}_\text{base}$   &109M  &22.5G \\
			
		     $\text{ViTCAP}$, $\text{BLIP}_\text{B}$ & $\text{ViT}_\text{B}$  &86.4M &55.5G   & $\text{BERT}_\text{base}$  &109M  &22.5G  \\
			
			 $\text{DistillVLM}$, MiniVLM  &Eff-DET  &7.5M  &4.4G  &MiniLM   &33M  &8.3G  \\
			
			 {\bf LightCap (Ours)}  & ResNet50  &23.5M  &4.1G  &$ \text{TinyBERT}_{4}$  &14.5M  &1.2G \\
			\hline
		\end{tabular*}
		\caption{Comparison of different captioning methods in terms of the model structure, inference speed in FLOPs (in G), number of parameters (in M).} \label{table:params and FLOP}	
	\end{center}
\end{table}

To the best of our knowledge, DistillVLM \cite{DistillVLM} and MiniVLM \cite{miniVLM} are the representative lightweight image captioners in the literature.
These methods design a tiny object detector called Eff-DET based on the EfficientNet \cite{Efficientnet}. 
Nevertheless, their fusion model (\emph{i.e.,} MiniLM \cite{MiniLM}) is still much larger than our $\text{TinyBERT}_4$.
As discussed in MiniVLM, changing the fusion model from MiniLM to a $\text{TinyBERT}_{4}$, the captioning performance will drop sharply (about 10 CIDEr).
Thanks to our designed concept extractor, cross-modal modulator, and ensemble head, a lightweight $\text{TinyBERT}_{4}$ also works well in our framework.
%

\setlength{\tabcolsep}{2pt}
\begin{table}
	\scriptsize
	\begin{center}
		\begin{tabular*}{8.5 cm} {@{\extracolsep{\fill}}l|cccc|cccc}
			\hline
			\multirow{2}{*}{Method}  & \multicolumn{4}{c|}{Cross-Entropy} & \multicolumn{4}{c}{CIDEr Optimization} \\
			  &B@4 &M &C &S &B@4 &M &C &S \\
			
			\hline
			 {\tiny \bf w/o Pre-training}  &  &  & & & &  & & \\
			
			
			
			 BUTD \cite{BUTD}   &36.2 &27.0 &113.5 &20.3  &36.3 &27.7 &120.1  &21.4\\
			 LBPF \cite{LBPF}   &37.4 &28.1 &116.4 &21.2  &38.3 &28.5 &127.6  &22.0\\
			 AoANet \cite{AoANet}   &37.2 &28.4 &119.8 &21.3  &38.9 &29.2 &129.8  &22.4\\
			
			
			 X-LAN \cite{XLAN}  &38.2 &28.8 &122.0 &21.9  &39.5 &29.5 &132.0  &23.4\\
			
			 RSTNet \cite{RSTNet}  &- &- &- &-  &39.3 &29.4 &133.3 &23.0\\
			 DLCT \cite{DLCT_AAAI}  &- &- &- &-  &39.8 &29.5 &133.8 &23.0\\
			
			\hline
			 {\tiny \bf Normal model design}  &  &  & & & &  & &\\
			 $\text{VLP}_\text{B}$ \cite{VLP}  &36.5 &28.4 &116.9 &21.2  &39.5 &29.3 &129.3  &23.2\\
			
			 $\text{Oscar}_\text{B}$ \cite{OSCAR}  &36.5 &30.3 &123.7 &23.1  &40.5 &29.7 &137.6 &22.8\\
			
			 $\text{UNIMO}_\text{B}$ \cite{UNIMO} &38.8 &- &124.4 &-  &- &- &- &-\\
			
			 $\text{ViTCAP}$ \cite{ViTCap}  &36.3 &29.3 &125.2 &22.6  &41.2 &30.1 &138.1 &24.1\\
			
			 $\text{VinVL}_\text{B}$ \cite{VINVL} &38.2 &30.3 &129.3 &23.6  &40.9 &30.9 &140.4 &25.1\\
			
			 $\text{LEMON}_\text{B}$ \cite{LEMON} &40.3 &30.2 &133.3 &23.3  &41.6 &31.0 &142.7 &25.1\\
			
			 $\text{BLIP}_\text{B}$ \cite{BLIP}  &39.7 &- &133.3 &23.3  &- &- &- &-\\
			
			
			\hline
			
			 {\tiny \bf Light model design}  &  &  &  & & &  & & \\
			 E2E-VLP \cite{E2E-VLP} &36.2 &- &117.3 &-  &- &- &- &-\\
			 $\text{MiniVLM}$  \cite{miniVLM} &35.6 &28.6 &119.8 &21.6  &39.2 &29.7 &131.7 &23.5\\
			 $\text{DistillVLM}$ \cite{DistillVLM}  &35.6 &28.7 &120.8 &22.1  &- &- &- &-\\
			
			 $\text{\bf LightCap (Ours)}$ &37.4  &29.9 &125.8 &22.6 &40.1 &29.9 &136.6 &24.2\\
			\hline
		\end{tabular*}
		\caption{Performance comparisons on the COCO Karpathy test split \cite{MSCOCO}.} \label{table:SOTA on COCO}	
	\end{center}
\end{table}

{\bf \noindent Evaluation on COCO.} In Table~\ref{table:SOTA on COCO}, we present the performance of state-of-the-art captioning methods on the COCO Karpathy test split \cite{karpathy}.
These approaches are generally trained with the cross-entropy loss and further optimized with CIDEr as a reinforcement learning reward.
Previous captioners without model pre-training such as BUTD, AoANet, and X-LAN mostly use the Faster R-CNN as the visual feature extractor.
The proposed LightCap outperforms all previous pretraining-free algorithms.

Recent ``pre-training then fine-tuning'' methods typically choose the BERT model as the cross-modal fusion model.
These methods struggle to achieve a fast inference speed with the large visual backbone and the heavyweight BERT model.
Using similar pre-training data and the same cross-entropy optimization, our LightCap (125.8 CIDEr) is superior to the heavyweight $\text{Oscar}_{\text{B}}$ (123.7 CIDEr) and $\text{UNIMO}_{\text{B}}$ (124.4 CIDEr).
Compared with other lightweight captioning methods such as MiniVLM and DistillVLM, our LightCap retains fewer parameters and FLOPs, but surpasses them by a notable margin of about 5 CIDEr score.
Note that BLIP and LEMON algorithms collect large-scale high-quality pre-training datasets containing 129 and 200 million image-text pairs (more than 20$\times$ larger than ours) for pre-training, respectively. 
%
%
We believe that the proposed LightCap can be further improved by involving more pre-training data, which leaves as our future work.

\setlength{\tabcolsep}{2pt}
\begin{table}
	\scriptsize
	\begin{center}
		\begin{tabular*}{8.3 cm} {@{\extracolsep{\fill}}l|cc|cc}
			\hline
			~ \multirow{2}{*}{Method}  & \multicolumn{2}{c|}{Out-of-domain}  & \multicolumn{2}{c}{Overall} \\
			~  &C &S &C &S \\
			
			\hline
			
			~ BUTD \cite{BUTD}  &31.3 &8.3 &55.3 &10.1\\
			
			~ BUTD \cite{BUTD} + CBS     &66.4 &9.7 &73.1 &11.1\\
			
			~ $\text{Oscar}_\text{B}$ \cite{OSCAR}   &45.3 &9.7 &63.8 &11.2 \\
			
			~ $\text{Oscar}_\text{B}$ \cite{OSCAR} + CBS   &77.6 &10.6 &81.1 &11.7 \\
			
			~ $\text{VIVO}_\text{B}$ \cite{VIVO}  &71.1 &10.6 &81.5 &12.2 \\
			
			~ $\text{VIVO}_\text{B}$ \cite{VIVO} + CBS   &87.5 &11.5 &88.3 &12.4 \\
			
			~ $\text{VinVL}_\text{B}$ \cite{VINVL} + CBS   &87.4 &11.6 &90.9 &12.8 \\
			
			~ $\text{ViTCAP}$ \cite{ViTCap}  &78.1 &11.9 &89.2 &12.7 \\
			
			~ $\text{ViTCAP}$ \cite{ViTCap} + CBS   &95.4 &12.7 &93.8 &13.0 \\
			
			~ \color{lightgray} $ \text{SimVLM}_\text{B}$ \cite{SimVLM} (w/ pre-train)  &\color{lightgray} - &\color{lightgray} - &\color{lightgray} 94.8 &\color{lightgray} 13.1 \\
			
			~ $\text{LEMON}_\text{B}$ \cite{LEMON}  &62.6 &10.6 &79.0 &12.3 \\
			~ \color{lightgray} $\text{LEMON}_\text{B}$ \cite{LEMON} (w/ pre-train) &\color{lightgray} 107.9 &\color{lightgray} 13.1 &\color{lightgray} 106.8 &\color{lightgray} 14.1 \\
			
			~ \color{lightgray}$\text{BLIP}_\text{B}$ \cite{BLIP} (w/ pre-train) &\color{lightgray} 111.5 &\color{lightgray} 14.2 &\color{lightgray} 109.6 &\color{lightgray} 14.7 \\
			
			\hline
			~ Human Performance   &95.7 &14.0 &87.1 &14.2 \\
			\hline
			
			~ $\text{\bf LightCap (Ours)}$  &76.5 &11.2 &85.1 &12.3\\
			
			~ $\text{\bf LightCap (Ours)}$ {\bf + CBS}  &90.5 &11.5 &90.8 &12.8\\
			
			\hline
		\end{tabular*}
		\caption{Performance comparisons on the nocaps validation split \cite{nocaps}. We report the results of both without and with constrained beam search (CBS) decoding.} \label{table:SOTA on nocaps}
	\end{center}
\end{table}

{\bf \noindent Evaluation on Nocaps.} 
Nocaps benchmark \cite{nocaps} contains 15,100 images collected from OpenImages \cite{OpenImage}. 
%
%
%
We evaluate the proposed method on the nocaps dataset to assess the model generalizability.
Due to the limited space, we only present the out-of-domain and overall performance in Table~\ref{table:SOTA on nocaps}.
%
%
Following the protocol of this benchmark, we merely train the LightCap model on the COCO-caption \emph{without} additional pre-training.
%
%
%
%
Our captioning model is much smaller than all the comparison methods such as VIVO and ViTCap.
%
%
It is also worth mentioning that our method surpasses the human CIDEr score and even slightly outperforms the strong VinVL method in the out-of-domain, which can be largely contributed to the representational power of the CLIP feature and our designed concept extractor to retrieve novel concepts.

\section{Conclusion}
In this paper, we propose a lightweight image captioning approach for resource-limited devices.
%
%
To unveil the potential of a capacity-limited tiny model, we design a visual concept extractor, a cross-modal modulator, and an ensemble head to improve the captioning quality.
%
%
By virtue of the sequential knowledge distillation and ensemble distillation, our LightCap exhibits competitive performance under a limited model capacity.
%
%
%
Extensive experiments verify the super-balanced performance and efficiency of the proposed LightCap.

{	\small
	\bibliographystyle{aaai}
	\bibliography{egbib}
}

\clearpage

\appendix

\section{Implementation Details}

\subsection{Training Details}

In the pre-training stage, the randomly initialized teacher models ($\text{BERT}_{\text{base}}$ \cite{BERT}) are trained $1.0$M steps with a learning rate of $5\times10^{-5}$ and batch size of $512$.
We use AdamW optimizer \cite{adam} with $\beta_{1}=0.9$, $\beta_{2} = 0.999$, and weight decay of $1\times10^{-2}$ to train the teacher models. 
Then, we leverage the pre-trained teacher models to jointly distill the student model in the same training settings, \emph{e.g.,} $1.0$M steps, learning rate $5\times 10^{-5}$ and batch size 512.
In the fine-tuning stage, the teacher models are trained $120$ epochs with a learning rate of $3\times10^{-5}$.
We empirically test 1, 2, 3, and 5 heads. We observe that two heads can obviously outperform a single head, but the performance tends to be saturated after 3 heads. Thus, we empirically set the head number to 3.
We utilize three strong teacher models to distill the student model using the same settings (\emph{e.g.,} AdamW optimizer, learning rate, and batch size).
The temperature $\tau$ in the KD process is set to 1.
The knowledge distillation on attentions and hidden states is conducted for 60 epochs, and the distillation on token probability is conducted for another 60 epochs.
Instead of ``training then distillation'', in the training stage, we combine the training loss and distillation loss to jointly train and distill the student model to save the training cost.

As for the visual concept number, we empirically set $K=20$ to select top-$K$ concepts for efficient cross-modal fusion. We observe that the performance will slightly drop when the concept number is less than 15.
Our visual concept extractor is trained on the VG dataset \cite{VG}, which is widely used in the image captioning task.

\subsection{Evaluation on the Mobile Device}

In this work, we test the inference latency of LightCap model on the mobile phone Huawei P40.
The testing chip on Huawei P40 mobile phone is Kirin 990\footnote{\url{https://www.hisilicon.com/en/products/Kirin/Kirin-flagship-chips/Kirin-990}}.
The detailed inference speeds of the components in LightCap are shown in Table~\ref{table:inference speed}.
To purely investigate the model inference speed, we set the beam search size to 1.
The memory usage is 257~MB on the mobile phone.
It merely takes about 188ms for our light model to process a single image on the CPU from mobile devices, which meets the real-world efficiency requirements.
It is well recognized that leveraging the NPU or GPU on mobile devices can achieve a higher inference speed, while not all the mobile devices are equipped with a strong chip.
Consequently, we utilize the CPU in Kirin 990 to test our method (188ms per image). 
The inference latency on the PC with a Titan X GPU is about 90ms.

\section{Visualization Results}

\subsection{Visualization of Visual Concept Extractor}
\begin{figure}[h]
	\centering
	\includegraphics[width=8.6cm]{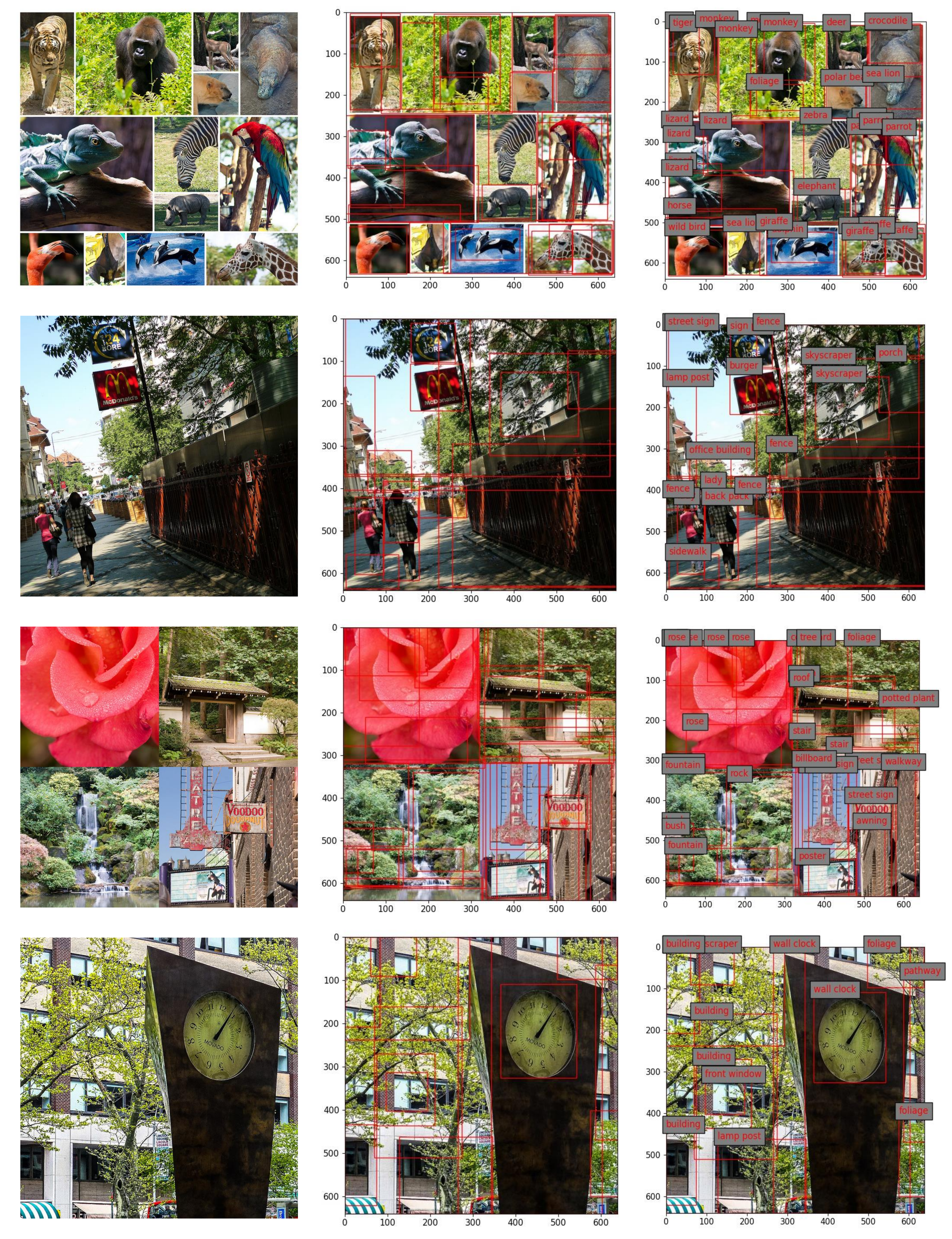}
	\caption{From left to right: input image, foreground detection results, and concept retrieval results. All the testing images are from COCO dataset \cite{MSCOCO}.}
	\label{fig:tag_visualization}
\end{figure}

We visualize the image concept retrieval results in Figure~\ref{fig:tag_visualization}.
In the second column, we exhibit the foreground detection results of the tiny detector YOLOv5n.
Although this detector is relatively weak and fails to outperform the state-of-the-art two-stage detection methods, it is extremely light with only 1.9M parameters.
Besides, accurate bounding boxes are not necessary for our framework.
Based on the roughly predicted foreground ROIs, we focus on retrieving visual concepts of the image.
As shown in the third column, our visual concept extractor is able to predict accurate and dense object tags to form the image concept.

\setlength{\tabcolsep}{2pt}
\begin{table}
	\scriptsize
	\begin{center}
		\caption{Inference latency of the proposed LightCap on the CPU device.} \label{table:inference speed}	
		\begin{tabular*}{8.4 cm} {@{\extracolsep{\fill}}l|cccc|c}
			\hline
			~ &Image  &Concept   &BERT Encoding  &BERT Decoding  &\multirow{2}{*}{Total}\\
			~ &Encoding &Extraction &(Img+Concept)  &(Caption)  &\\
			\hline 
			~  Time &110.1ms  &20.0ms  &11.4ms  &3.8ms$\times$12 (caption length)   & 188.3ms   \\
			\hline
		\end{tabular*}
	\end{center}
\end{table}

\begin{figure*}[h]
	\centering
	\includegraphics[width=14.0cm]{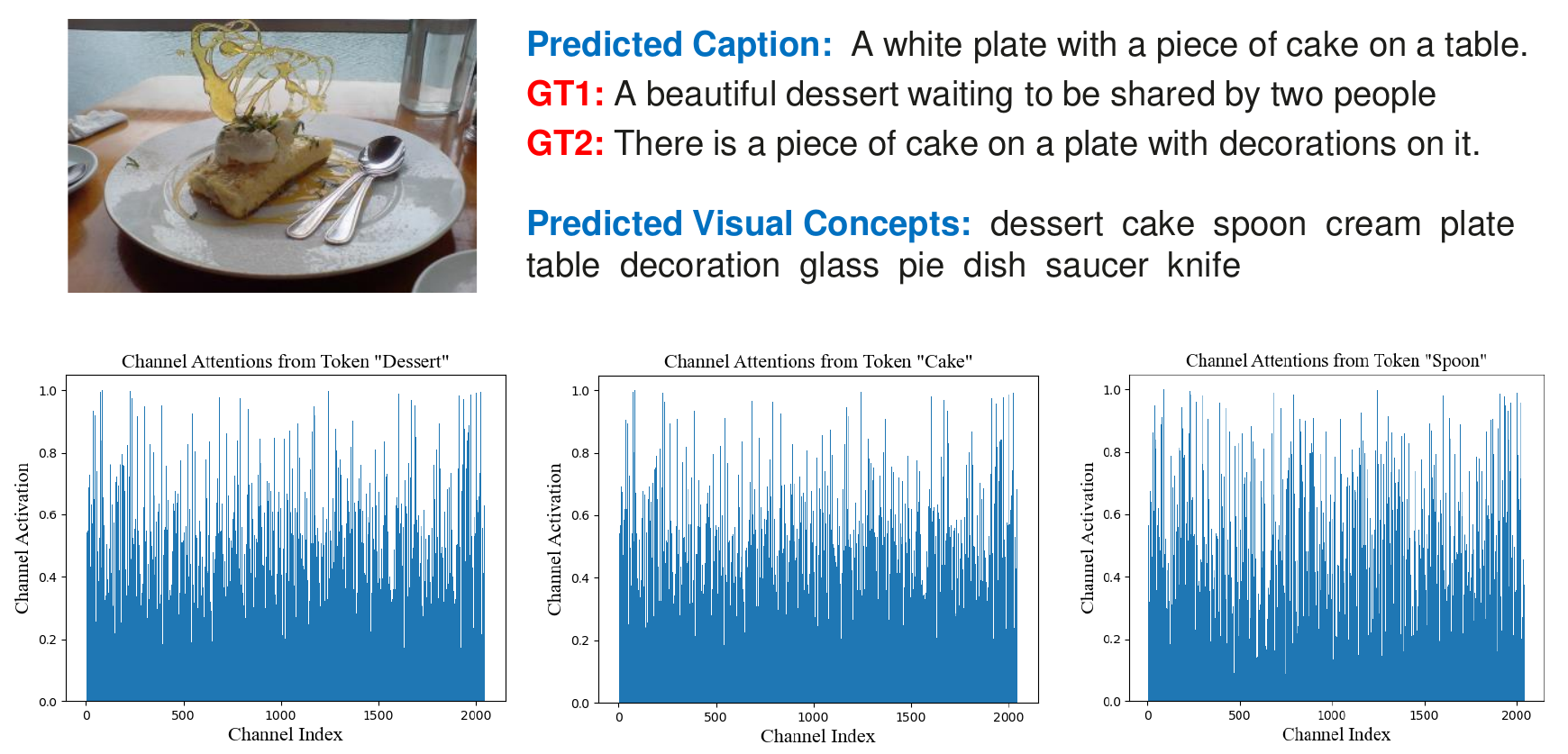}
	\caption{In the top figure, we show the predicted image caption, ground truth (GT) captions, and our predicted visual concepts. In the bottom figure, we exhibit the channel attention weights of the first three concepts (\emph{i.e.,} {\tt Dessert}, {\tt Cake}, and {\tt Spoon}). }
	\label{fig:channel_attention}
\end{figure*}

\begin{figure*}[h]
	\centering
	\includegraphics[width=16.2cm]{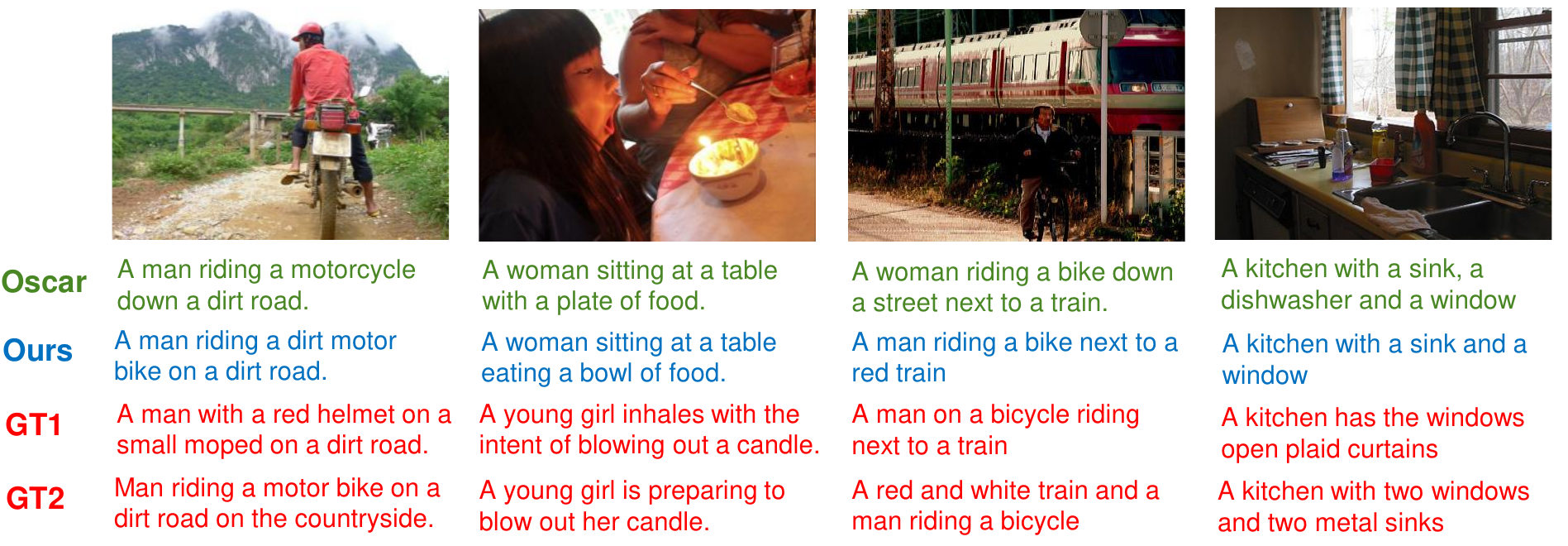}
	\caption{ Uncurated image captioning examples of the first four images in COCO Karpathy test split \cite{karpathy}, coupled with the correspondence ground truth (GT) sentences. }
	\label{fig:example}
\end{figure*}

\subsection{Visualization of Cross-modal Modulator}

In Figure~\ref{fig:channel_attention}, we further visualize the channel attentions of the retrieved visual concepts.
For the given image in Figure~\ref{fig:channel_attention}, the first three visual concepts are {\tt Dessert}, {\tt Cake}, and {\tt Spoon}.
These visual concepts are projected to the channel attentions to modulate the raw CLIP features.
As shown in the bottom figures in Figure~\ref{fig:channel_attention}, the activated channels are sparse (\emph{i.e.,} only a few channels yield the high attention values of more than 0.8) and most channel weights are below 0.5.
This verifies our assumption that the raw CLIP features are redundant in the channel dimension.
Besides, the channel attentions from {\tt Dessert} and {\tt Cake} are similar, potentially due to their high similarity in the semantic space.
However, the attention weight generated by {\tt Spoon} is quite different from the attentions of {\tt Dessert} and {\tt Cake}.
It is well recognized that different feature channels represent certain semantics, and our approach is able to activate the informative channels using the retrieved concepts for effective image captioning.

\subsection{Qualitative Evaluation} 
Finally, we exhibit the captioning results of our approach on the COCO-caption dataset \cite{karpathy} in Figure~\ref{fig:example}, coupled with ground truth (GT) sentences.
Figure~\ref{fig:example} also showcases the results of the state-of-the-art $\text{Oscar}_{\text{B}}$ method \cite{OSCAR}.
Overall, on these uncurated images from the COCO Karpathy test set, our LightCap generates accurate captions and is comparable with the strong $\text{Oscar}_{\text{B}}$.
The proposed approach even yields more accurate captions than $\text{Oscar}_{\text{B}}$ in the third picture, where $\text{Oscar}_{\text{B}}$ predicts \texttt{woman} instead of \texttt{man}.
It should be noted that such a robust model achieves promising results by retaining only 2\% FLOPs of the current state-of-the-art captioners.

\section{Results on Nocaps}

Due to the limited space, we only exhibit ``out-of-domain'' and ``overall'' comparison  results on the Nocaps dataset \cite{nocaps} in the main paper.
In Table~\ref{table:SOTA on nocaps supple} of this supplementary material, we show the complete results including ``in-domain'', ``near-domain'', ``out-of-domain'', and ``overall'' performance.

\setlength{\tabcolsep}{2pt}
\begin{table*}[h]
	\scriptsize
	\begin{center}
		\begin{tabular*}{13.0 cm} {@{\extracolsep{\fill}}l|cc|cc|cc|cc}
			\hline
			~ \multirow{2}{*}{Method}  & \multicolumn{2}{c|}{In-domain~~~~} & \multicolumn{2}{c|}{Near-domain} & \multicolumn{2}{c|}{Out-of-domain}  & \multicolumn{2}{c}{Overall} \\
			~  &C &S &C &S &C &S &C &S \\
			
			\hline
			
			~ BUTD \cite{BUTD}  &78.1 &11.6 &57.7 &10.3 &31.3 &8.3 &55.3 &10.1\\
			
			~ BUTD \cite{BUTD} + CBS    &80.0 &12.0 &73.6 &11.3 &66.4 &9.7 &73.1 &11.1\\
			
			~ $\text{Oscar}_\text{B}$ \cite{OSCAR}    &79.6 &12.3 &66.1 &11.5 &45.3 &9.7 &63.8 &11.2 \\
			
			~ $\text{Oscar}_\text{B}$ \cite{OSCAR} + CBS   &83.4 &12.0 &81.6 &12.0 &77.6 &10.6 &81.1 &11.7 \\
			
			~ $\text{VIVO}_\text{B}$ \cite{VIVO}  &88.8 &12.9 &83.2 &12.6 &71.1 &10.6 &81.5 &12.2 \\
			
			~ $\text{VIVO}_\text{B}$ \cite{VIVO} + CBS  &92.2 &12.9 &87.8 &12.6 &87.5 &11.5 &88.3 &12.4 \\
			
			~ $\text{VinVL}_\text{B}$ \cite{VINVL} + CBS  &96.8 &13.5 &90.7 &13.1 &87.4 &11.6 &90.9 &12.8 \\
			
			~ $\text{ViTCAP}$ \cite{ViTCap}  &99.3 &13.2 &90.4 &12.9 &78.1 &11.9 &89.2 &12.7 \\
			
			~ $\text{ViTCAP}$ \cite{ViTCap} + CBS  &98.7 &13.3 &92.3 &13.3 &95.4 &12.7 &93.8 &13.0 \\
			
			~ \color{lightgray} $ \text{SimVLM}_\text{B}$ \cite{SimVLM} (w/ pre-train)  &\color{lightgray} - &\color{lightgray} - &\color{lightgray} - &\color{lightgray} - &\color{lightgray} - &\color{lightgray} - &\color{lightgray} 94.8 &\color{lightgray} 13.1 \\
			
			~ $\text{LEMON}_\text{B}$ \cite{LEMON}  &91.4 &13.3 &81.4 &12.5 &62.6 &10.6 &79.0 &12.3 \\
			~ \color{lightgray} $\text{LEMON}_\text{B}$ \cite{LEMON} (w/ pre-train) &\color{lightgray} 107.7 &\color{lightgray} 14.7 &\color{lightgray} 106.2 &\color{lightgray} 14.3 &\color{lightgray} 107.9 &\color{lightgray} 13.1 &\color{lightgray} 106.8 &\color{lightgray} 14.1 \\
			
			~ \color{lightgray}$\text{BLIP}_\text{B}$ \cite{BLIP} (w/ pre-train) &\color{lightgray} 111.8 &\color{lightgray} 14.9 &\color{lightgray} 108.6 &\color{lightgray} 14.8 &\color{lightgray} 111.5 &\color{lightgray} 14.2 &\color{lightgray} 109.6 &\color{lightgray} 14.7 \\
			
			\hline
			~ Human Performance   &84.4 &14.3 &85.0 &14.3 &95.7 &14.0 &87.1 &14.2 \\
			\hline
			~ $\text{\bf LightCap (Ours)}$  &95.4 &13.2 &85.5 &12.3 &76.5 &11.2 &85.1 &12.3\\
			~ $\text{\bf LightCap (Ours)}$ {\bf + CBS} &95.8 &13.4 &88.7 &12.8 &90.5 &11.5 &90.8 &12.8\\
			
			\hline
		\end{tabular*}
		\caption{Performance comparisons on the Nocaps validation split \cite{nocaps}, where C and S denote CIDEr and SPICE scores. We compare our method with previous state-of-the-art approaches at ``in-domain", ``near-domain", and ``out-of-domain". We report the results of both without and with constrained beam search (CBS) decoding.} \label{table:SOTA on nocaps supple}	
	\end{center}
\end{table*}

\setlength{\tabcolsep}{2pt}
\begin{table*}
	\scriptsize
	\begin{center}
		\begin{tabular*}{13.0 cm} {@{\extracolsep{\fill}}l|cc|c|cccc}
			\hline
			~ \multirow{2}{*}{Method} & \multicolumn{2}{c|}{Model Architecture}  &Pre-training & \multicolumn{4}{c}{Cross-Entropy}  \\
			~ & Image Encoder & Fusion Model &Data  &B@4 &M &C &S \\
			
			\hline
			
			~ {\tiny \bf Normal model design}  &  &  & & & & &\\
			
			~ $\text{VLP}_\text{B}$ \cite{VLP}  & $\text{F-RCNN}_\text{101}$ & $\text{BERT}_\text{base}$ &4M &36.5 &28.4 &116.9 &21.2  \\
			
			~ $\text{Oscar}_\text{B}$ \cite{OSCAR}  & $\text{F-RCNN}_\text{101}$ & $\text{BERT}_\text{base}$ &7M &36.5 &30.3 &123.7 &23.1 \\
			
			~ $\text{UNIMO}_\text{B}$ \cite{UNIMO}  & $\text{F-RCNN}_\text{101}$ & $\text{BERT}_\text{base}$ &9M &38.8 &- &124.4 &- \\
			
			~ $\text{ViTCAP}$ \cite{ViTCap} & $\text{ViT}_\text{B}$ & $\text{BERT}_\text{base}$ &10M &36.3 &29.3 &125.2 &22.6  \\
			
			~ $\text{VinVL}_\text{B}$ \cite{VINVL}  & $\text{ResNeXt}_\text{152}$ & $\text{BERT}_\text{base}$ &9M &38.2 &30.3 &129.3 &23.6 \\
			
			~ $\text{LEMON}_\text{B}$ \cite{LEMON}  & $\text{ResNeXt}_\text{152}$ & $\text{BERT}_\text{base}$ &200M &40.3 &30.2 &133.3 &23.3 \\
			
			~ $\text{BLIP}_\text{B}$ \cite{BLIP}  & $\text{ViT}_\text{B}$ & $\text{BERT}_\text{base}$ &129M &39.7 &- &133.3 &23.3 \\
			
			~ $\text{SimVLM}_\text{B}$ \cite{SimVLM}  & $\text{ResNet}$\&$\text{ViT}_\text{B}$  & $\text{Transformer}$   &1.8B &39.0 &32.9 &134.8  &24.0 \\
			\hline
			~ {\tiny \bf Light model design}  &  &  &  & & & & \\
			~ E2E-VLP \cite{E2E-VLP} &$\text{ResNet}_\text{50}$    & Transformer  &6M &36.2 &- &117.3 &-  \\
			~ $\text{MiniVLM}$  \cite{miniVLM} &Eff-DET    & MiniLM  &14M &35.6 &28.6 &119.8 &21.6  \\
			~ $\text{DistillVLM}$ \cite{DistillVLM}  &Eff-DET & MiniLM  &7M &35.6 &28.7 &120.8 &22.1  \\
			~ $\text{\bf LightCap (Ours)}$   &$\text{ResNet}_\text{50}$ & $\text{TinyBERT}_\text{4}$ &6M &37.4  &29.9 &125.8 &22.6 \\
			\hline
		\end{tabular*}
		\caption{ Performance comparisons on the COCO-caption Karpathy test split \cite{MSCOCO}, where B@4, M, C, S denote BLEU@4, METEOR, CIDEr, and SPICE scores.} \label{table:SOTA on COCO supp}	
	\end{center}
\end{table*}

\section{Limitations and Future Work}
Despite the super-balanced performance and efficiency, the proposed framework still has some limitations:

{\flushleft \bf (1) Training a More Efficient CLIP.} The main computational cost of our work lies in the visual backbone (\emph{i.e.,} ResNet-50). In the future, we plan to train an EfficientNet-based CLIP model to further reduce the feature extraction latency of the visual encoder. 

{\flushleft \bf (2) End-to-end Training.} Currently, we freeze the model parameters of the CLIP ResNet-50 backbone. We observe that end-to-end training of the visual backbone will degrade the performance, potentially due to the limited training data in the image captioning domain.
In the future, we intend to include more data to facilitate the joint training of the visual backbone and fusion model.

{\flushleft \bf (3) Adding More Pre-training Data.} Although our approach adopts the cross-modal pre-training, as shown in Table~\ref{table:SOTA on COCO supp}, our pre-training data is much less than the recent LEMON \cite{LEMON}, BLIP \cite{BLIP}, and SimVLM \cite{SimVLM}. In the future, we plan to involve more pre-training data to boost the captioning quality.

\end{document}